%% file: autononovi_arxiv_main.tex
\documentclass[letterpaper, 10 pt, conference]{ieeeconf}  

\IEEEoverridecommandlockouts                              

\usepackage{color}
\usepackage{amsmath}
\usepackage{amssymb}
\usepackage{graphicx}
\usepackage{wrapfig}
\usepackage{bm} 

\newcommand*{\secref}[1]{section \ref{#1}}
\newcommand*{\eqnref}[1]{equation (\ref{#1})}
\newcommand*{\figref}[1]{figure \ref{#1}}

\newcommand*{\Figref}[1]{Figure \ref{#1}}

\newcommand*{\mc}[1]{\mathcal{#1}}

\let\oldvec\vec
\renewcommand{\vec}[1]{\boldsymbol{\oldvec{#1}}}

\title{\LARGE \bf
AutonoVi: Autonomous Vehicle Planning with Dynamic Maneuvers and Traffic Constraints
}

\begin{document}

\author{Andrew Best$^{1}$ and Sahil Narang$^{1}$ 
and Daniel Barber$^{2} $ and Dinesh Manocha$^{1}$ \\
http://gamma.cs.unc.edu/AutonoVi/video.avi  (video included)
\thanks{$^{1}$Andrew Best and Sahil Narang and Dinesh Manocha are at the University of North Carolina, Chapel Hill}
\thanks{$^{2}$ Daniel Barber is at the University of Central Florida}
}



\maketitle
\thispagestyle{empty}
\pagestyle{empty}

\input{abstract}
\input{introduction}
\input{related_work}
\input{problem_space}

\input{algorithm}
\input{experiment_definitions}

\input{experiment_results}
\input{conclusion}

\vspace*{-0.09in}

\bibliographystyle{unsrt}
\bibliography{auto_driving}

\end{document}

%% file: abstract.tex
\begin{abstract}

We present AutonoVi:, a novel algorithm for autonomous vehicle navigation that supports dynamic maneuvers and satisfies traffic constraints and norms.
Our approach is based on optimization-based maneuver planning that supports dynamic lane-changes, swerving, and braking in all traffic scenarios and guides the vehicle to its goal position. We take into account various traffic constraints, including collision avoidance with other vehicles, pedestrians, and cyclists using control velocity obstacles. We use a data-driven approach to model the vehicle dynamics for control and collision avoidance. Furthermore, our trajectory computation algorithm takes into account traffic rules  and behaviors, such as stopping at intersections and stoplights, based on an arc-spline representation. We have evaluated our algorithm in a simulated environment and tested its interactive performance in urban and highway driving scenarios with tens of vehicles, pedestrians, and cyclists. These scenarios include jaywalking pedestrians, sudden stops from high speeds, safely passing cyclists, a vehicle suddenly swerving into the roadway, and high-density traffic where the vehicle must change lanes to progress more effectively.

\end{abstract}




%% file: introduction.tex
\section{INTRODUCTION}

Autonomous driving is a difficult and extremely complex task that has immense potential for impacting the lives of billions of people. In order to develop autonomous capabilities to perform the driving task, we need appropriate capabilities to sense and predict the traffic and road obstacles, as well as for planning, control, and coordination of the vehicle~\cite{Ziegler2014b,Pendleton2017}. There is considerable research in this area that borrows ideas from different disciplines including computer vision, machine learning, motion planning, mechanical engineering, intelligent traffic simulation, human-factors psychology, etc.

Research into sensing and perception technologies has been progressing considerably and current vehicle sensors seem to have the capability to detect relevant obstacles, vehicles, and other traffic entities including bicycles and pedestrians. However, automatic planning in different scenarios and the computation of the appropriate response to vehicle and non-vehicle entities, such as bicycles and pedestrians, are still the subjects of ongoing research. A key issue is the development of an efficient navigation algorithm for autonomous driving that takes into account the vehicle dynamics, sensor inputs, traffic rules and norms, and the driving behaviors of other vehicles.
Moreover, the uncertainties in the sensor data, capability, and response of the autonomous vehicle, typically referred to as the {\em ego-vehicle} \cite{Ziegler2014}, have led to the development of behavior and navigation algorithms that impose conservative limits on the acceleration, deceleration, and steering decisions. For example, algorithms tend to limit hazard responses to either steering \cite{Borrelli2005}, \cite{Eidehall2007} or braking \cite{Distner2009}. Few algorithms demonstrate combined control of throttle and steering and typically do so in constrained navigation scenarios \cite{Turri2013}. 
In terms of planning the routes and navigating the roads, current algorithms tend to limit the lane-changing behaviors, precluding their use for progressing more quickly to a goal or better utilization of the road conditions.
These limitations have led to the perception that autonomous cars behave more like student drivers taking their driving test than actual skilled human drivers~\cite{Ziegler2014}. One of the goals is to extend the capabilities of current autonomous vehicles in terms of planning, control, and navigation, making them less conservative but still allowing safe performance during driving.


\textbf{Main contributions}:  We present a novel navigation algorithm for autonomous vehicles, AutonoVi, which utilizes a data-driven vehicle dynamics model and optimization-based maneuver planning to compute a safe, collision-free trajectory with  dynamic lane-changes. Our approach is general, makes no assumption about the traffic conditions, and  plans dynamically feasible maneuvers  in traffic consisting of other vehicles, cyclists, and pedestrians. In order to develop an autonomous vehicle planning approach with these capabilities, we present four novel algorithms: 

\begin{itemize}
    \item {\bf Optimization-based Maneuvering:} We describe a novel multi-objective optimization approach for evaluating the dynamic maneuvers.  Our algorithm encodes passenger comfort, safe passing distances, maneuver constraints in terms of dynamics, and global route progress in order to compute appropriate trajectories. 
     
     \item {\bf  Data driven Vehicle Dynamics:} We use a data-driven vehicle dynamics formulation that encodes feasible accelerations, steering rates, and decelerations into a set of per-vehicle profile functions, which can be quickly evaluated. These profiles are generated by simulating the ego-vehicle through a series of trials to obtain lateral and longitudinal slip profiles. This data-driven model generalizes to multiple vehicles and configurations. 
    
    \item {\bf  Collision avoidance with kinematic and dynamic constraints:} We present a collision avoidance algorithm that combines collision-free constraints with specific kinematic and dynamic constraints of the autonomous vehicle. Our approach allows the autonomous vehicle  to steer away from collisions with other vehicles, pedestrians, and cyclists as well as to apply brakes, or use a combination of steering and braking.
    
    \item {\bf Trajectory Planning with Traffic Rules and Behaviors:} We present a trajectory planning algorithm that encodes traffic rules and road behaviors along with lane-following for computing safe trajectories. Our approach is based on computing arcs along the center-line of the current lane to generate an initial trajectory that satisfies all the constraints. 
    This initial trajectory is computed and refined according to collision avoidance and maneuver optimization computations.
   
\end{itemize}

We evaluate our algorithm in a set of traffic scenarios generated using a physics-based traffic simulator in both sparse and dense traffic conditions with tens of other vehicles, pedestrians, and cyclists. 
We demonstrate collision-avoidance events including a vehicle suddenly driving into the road, traffic suddenly stopping ahead of the ego-vehicle while travelling at high speed, and a pedestrian jaywalking in front of the ego-vehicle, representing typical accident scenarios \cite{Eidehall2007}. Our approach enables advantageous of lane changes (e.g., overtaking) and adherence to traffic rules in typical traffic conditions. It also exhibits safe maneuvering in the presence of heavy traffic, pedestrians, and cyclists. 
To our knowledge, AutonoVi is the first approach that allows the ego-vehicle to follow an arbitrary route, determine appropriate lane changes dynamically during travel, and plan dynamically and kinematically feasible trajectories while following traffic norms and providing collision avoidance for vehicles, pedestrians, and cyclists.

The rest of the paper is organized as follows: we detail relevant related work in section 2. In section 3, we introduce the vehicle kinematic model, define relevant assumptions, and introduce the terminology used in the rest of the paper. In section 4, we present our navigation algorithm, AutonoVi, and its components.
We present the details of our simulation benchmarks in \secref{sec:experiment:setup} and highlight the results in \secref{sec:experiment:results}.

%% file: related_work.tex
\section{RELATED WORK}

The problem of autonomous driving has been widely studied in robotics, computer vision, intelligent transportation systems and related areas. In this section, we give a brief overview of prior methods which address motion planning and navigation, dynamics, behavior generation, and collision avoidance. More detailed surveys are given in~\cite{Pendleton2017,Katrakazas2015,Saifuzzaman2014}.

\textbf{Vehicle Kinematics and Dynamics Modeling}: A number of approaches have been developed to model the motion of a moving vehicle. Different models offer a trade-off between simplicity or efficiency of the approach, and physical accuracy. Simpler models are typically based on linear dynamics and analytical solutions to the equations of motion~\cite{LaValle2006Book}. More accurate models provide a better representation of the physical motion, but require more computational power to evaluate and incorporate non-linear forces in the vehicle dynamics~\cite{Borrelli2005}.
The Reeds-Shepp formulation is a widely used car model with forward and backward gears~\cite{Reeds1990}. Margolis and Asgari~\cite{Margolis1991} present several representations of a car including the widely used single-track bicycle model. Borrelli et al.~\cite{Borrelli2005} extend this model by including detailed tire-forces. Current planning and control algorithms leverage varying levels of detail in the model of the vehicle. 

\textbf{Path Planning and Collision Avoidance}: 
Prior approaches to  path planning for autonomous vehicles are based on occupancy grids \cite{Kolski2006}, random-exploration~\cite{Kuwata2009}, driving corridors~\cite{Hardy2013}, potential-field methods~\cite{Galceran2015a}, etc. Recent approaches seek to incorporate driver behavior prediction in path planning using game-theoretic approaches~\cite{Sadigh2016} and Bayesian behavior modeling~\cite{Galceran2015}. In addition, a variety of algorithms have been proposed for planning paths for automobiles
for navigation outside of road conditions and traffic rules ~\cite{Ziegler2008}. Several techniques have been proposed to specifically avoid hazards while remaining in a target lane. These techniques can be coupled with a path planner to avoid vehicles \cite{Fritz2004} and other hazards in the ego-vehicle's lane \cite{Turri2013}.  

Many continuous approaches for collision-avoidance have been proposed based on spatial decomposition or velocity-space reasoning. Berg et al.~\cite{VanDenBerg2011a} apply velocity-space reasoning with acceleration constraints to generate safe and collision-free velocities. Bareiss et al.~\cite{Bareiss2015} extend the concept of velocity obstacles into the control space to generate a complete set of collision free control inputs. 
Ziegler et al.~\cite{Ziegler2014b} utilize polygonal decomposition of obstacles to generate blockages in continuous driving corridors. Sun et al.~\cite{Sun2014} demonstrate the use of prediction functions and trajectory set generation to plan safe lane-changes. 

\textbf{Modeling Traffic Rules}: Aside from planning the appropriate paths to avoid collisions, autonomous vehicles must also follow applicable laws and traffic norms. Techniques have been proposed to simulate typical traffic behaviors in traffic simulation such as Human Driver Model \cite{Treiber2006} and data-driven models such as \cite{Hidas2005}. An extensive discussion on techniques to model these behaviors in traffic simulation can be found in \cite{Chen2010}.


\begin{figure*}
    \centering
    \includegraphics{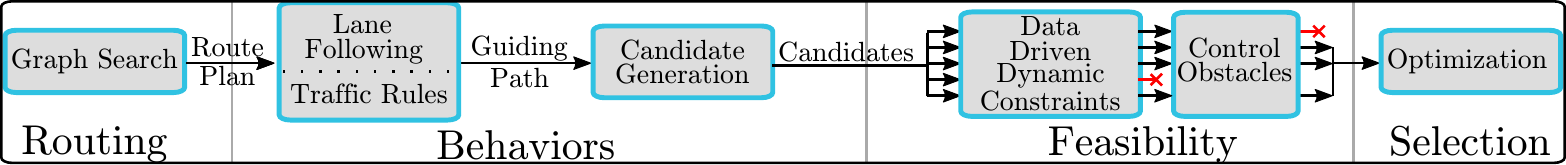}
    \caption{{\bf Algorithm Pipeline:} Our autonomous vehicle planning algorithm operates in several sequential steps. First, a route is planned using graph-search over the network of roads. Secondly, traffic and lane-following rules are combined to create a guiding trajectory for the vehicle for the next planning phase. This guiding trajectory is transformed to generate a set of candidate control inputs. These controls are evaluated for dynamic feasibility using our data driven vehicle dynamics modeling and collision-free navigation via extended control obstacles. Those remaining trajectories are  evaluated using our optimization technique to determine the most-appropriate set of controls for the next execution cycle.}
    \label{fig:pipeline}
    \vspace*{-0.1in}
\end{figure*}

\textbf{Autonomous Driving Systems}: 
Many autonomous systems have been demonstrated that are able to navigate an autonomous vehicle in traffic along a specific route. Ziegler et al.~\cite{Ziegler2014} demonstrated an autonomous vehicle which drove the historic Bertha Benz route in southern Germany. They use a conservative navigation approach, which specifically encodes {\em lanelets} for lane changing and does not account for dynamic lane changes. In contrast, our algorithm allows the vehicle to change lanes when our maneuver optimization method deems it appropriate and does not rely on pre-encoded changes. Geiger et al.~\cite{Geiger2012} demonstrate a planning and control framework that won the Grand Cooperative Driving Challenge in 2011. This vehicle was designed for platooning and employed controls over acceleration only. Our navigation algorithm plans maneuvers using both steering and acceleration to operate in more generic traffic scenarios. The DARPA Urban Grand Challenge included a number of autonomous vehicle navigating examples of driving scenarios~\cite{Urmson2009,Bacha2008}. While overtaking was allowed as an intended capability in these systems, the vehicles were not evaluated in dense, high-speed traffic conditions where the benefits of lane changes could be demonstrated. 

%% file: problem_space.tex
\section{Problem Space}

In this section, we introduce the notation, the kinematic and dynamics model of the car and the  state space of the vehicle in terms of both physical configuration and behavior space.



\subsection{Vehicle State Space}

We represent the kinematic and dynamic constraints of the vehicle separately. In terms of trajectory planning, steering and throttle controls that could lead to skidding or a loss of control are first excluded in our dynamics model (see \secref{subsec:profiling}) and future trajectories are computed according to our vehicle kinematic model described in \eqnref{eqn:motionequations}. 

We extend the simple-car kinematic model \cite{LaValle2006Book,Laumond98guidelinesin}. The vehicle has three degrees of freedom in a planar coordinate space. These are the position of the center of mass $\vec{p} = (p_x, p_y)$, and the current heading or orientation $\theta$. We represent the speed of the vehicle as $v$ and steering as $\phi$. $L_f$ and $L_r$ represent the distance from the center of mass to the front and rear axles, respectively. The geometry of the ego-vehicle is represented as $\mathcal{O}_e$.

The vehicle has two degrees of control, throttle ($u_t$) and steering ($u_\phi$). We define throttle $-1 \leq u_t \leq 1$, where $-1$ indicates maximum braking effort for the vehicle and $1$ represents maximum throttle. $-1 \leq\ u_\phi \leq  1$ describes the steering effort from $-\phi_{max}$ to $\phi_{max}$.  

We also use acceleration and steering functions, $A(v, u_t)$ and $\Phi(v, u_s)$, respectively, which describe the relationship between the vehicle's speed, steering, and control inputs and its potential for changes in the acceleration and steering (see \secref{subsec:profiling}). $A$ and $\Phi$ can be chosen to be constants in the simplest model, or may be represented using complex functions corresponding to tire dynamics and load transfer.
We describe our choice for $A$ and $\Phi$ in \secref{subsec:profiling}. The vehicle's motion can be described by: 

\begin{subequations}\label{eqn:motionequations}
\begin{align}
    \dot{p}_x &= v\cos(\theta) & \dot{p}_y &= v\sin(\theta) & \dot{\theta} &= \frac{\tan(\phi)}{L_f + L_r}v\label{eqn:motion:firstorder} \\
    \dot{v} &= A(v, u_t) & \dot{\phi} &= \Phi(\phi, u_s)  \label{eqn:motion:secondorder}
\end{align}
\end{subequations}

In addition to the physical state of the vehicle, we describe its behavior $b$ as a label from a set of all behaviors $\mc{B}$, such as driving straight, turning left, merging right, etc. The behavior state is used to modify parameters of each stage of the algorithm. Each behavior state can encode a set of weights of the maneuver optimization function, bias the generation of a guiding path, and adjust the sampling bias of our control-obstacle approximation and acceleration when necessary (see \secref{sec:behavior_state}). The full state of a vehicle is defined as $X_e = \{p_x, p_y, v, \phi, u_t, u_\phi, b\}$. 
The vehicle updates its plan at a fixed planning rate $\Delta t$. At each planning step, the vehicle computes a target speed $v'$ and target steering $\phi'$ to be achieved by the control system. We refer to \eqnref{eqn:motionequations} compactly as the \emph{state evolution function}  $X_{t+\Delta t} = q(X_t, u, t)$. We also define a function $S(u, X)$ which determines if a set of controls is feasible. Given the current state of the vehicle, $S(u, X)$ will return false if the given input $u$ will cause a loss of traction or control. We describe this function further in \secref{subsec:profiling}.

\subsection{Sensing and Perception}

We assume a sensing module is available for the vehicle that is capable of providing  information regarding the surrounding environment. For each lane on a road, the sensing module provides an approximation of the center line of the lane, $l$. The sensing module also provides the closest point on the lane center to the ego-vehicle, $\vec{l}_p$, and a reasonable value of the friction coefficient $\mu$. Recent work has presented approaches to evaluate $\mu$ from sensor data \cite{Gustafsson1997}. Our navigation algorithm utilizes the set of nearby vehicles, pedestrians, bicycles, or other obstacles, collectively referred to as neighbors, $N$ within the sensing range. For each neighbor $n \in N$, the sensing system provides the neighbor's shape, $\mathcal{O}_n$, position, $\vec{p}_n$, and velocity $\vec{v}_n$. Moreover, the sensing module provides the lane $l_n$, acceleration $\dot{v}_n$, and rate of turn, $\dot{\theta}$ for the neighbor. We define a set of neighbor types, $\mc{T}_n$, including vehicle, pedestrian, cyclist, and obstruction. Each neighbor is assigned a type $\mc{T}_n$ corresponding to the detected neighbor type. The complete state of a neighbor is denoted as $X_n = \{\vec{p}_n, \vec{v}_n, l_n, \dot{v}_n, \dot{\theta}_n\}$

%% file: algorithm.tex
\section{Navigation Algorithm}

In this section, we describe our navigation algorithm. Our algorithmic approach operates in four sequential stages, shown in Fig.~\ref{fig:pipeline}. First, a route is constructed over the space of roads in the environment. Secondly, a \emph{Guiding Path} that follows the current lane is computed that provides input to the collision-avoidance and optimization-based maneuver
stages. The collision avoidance stage determines the set of feasible \emph{candidate controls} that represent dynamically feasible, collision-free controls for the vehicle. Finally, a new control is chosen for the vehicle based on the optimization-based maneuver function. 

\subsection{Route Choice and Behavior State}
\label{sec:behavior_state}


\begin{figure}
  \begin{center}
    \includegraphics[]{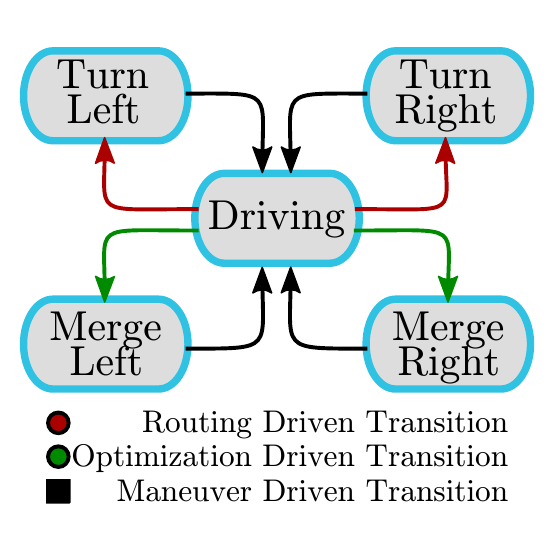}
  \end{center}
  \vspace{-16pt}
  \caption{ {\bf Finite State Machine:} We highlight different behavior states that are determined by the routing and optimization algorithms. When executing turns, the routing algorithm transitions the behavior state to a turning state. When the optimization-based maneuver algorithm plans a lane change, the behavior state is transitioned to merging. }
  \label{fig:fsm}
\end{figure}

Our navigation algorithm performs several steps in a sequential manner. In the first step, a global route for the vehicle to follow to the goal is determined. This step is performed only once unless special conditions (e.g.,  missing a turn) force the vehicle to recompute a route. The ego-vehicle is provided a connected graph of roads in the environment from a GIS database. Each road in the graph contains information on the number and configuration of lanes in the road and the speed limits. When a destination is chosen, we use A* search to compute the shortest route to the goal and construct a route plan. 

Each step of the route plan encodes how the vehicle transitions from one road to the next. We denote these as \emph{road-transition maneuvers}. A road-transition maneuver consists of the valid source lanes, valid destination lanes, the position along the road at which the maneuver begins, denoted $\vec{p}_m$, and the behavior implied by the road transition. The set of behaviors includes merging, right turns, left turns, and driving straight. Once the road-change maneuver is completed, the vehicle navigates along the lanes of the new road until the next maneuver node is reached. Lane changes are not encoded in the maneuver nodes, but they are performed implicitly based on the optimization function described in \secref{sec:optimization}.

The behavior state of the vehicle is described by a finite-state machine shown in \figref{fig:fsm}. It is used to restrict potential control decisions and adjust the weight of the cost-function for specific maneuvers, such as turning. This allows our algorithm to force the vehicle to be more conservative when performing delicate maneuvers. For example, the valid steering space is constrained in turns to guarantee that the vehicle moves closely along the center line. 

\subsection{Guiding Path}
\label{sec:guiding_trajectory}

\begin{figure}
    \centering
    \includegraphics[scale=0.8]{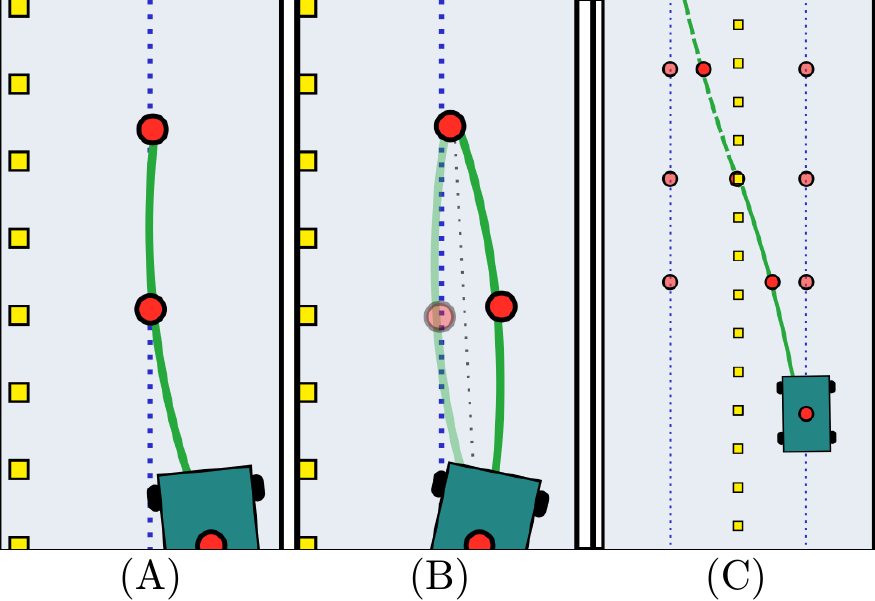}
    \caption{{\bf Guiding Path Computation:} The vehicle computes a guiding path to the center of its current lane based on a circular arc. {\bf (A):} When the vehicle tracks a path off the center of its current lane, the guiding path leads it smoothly back to center. {\bf (B):} In cases where the guiding path represents abrupt changes to heading, the center point is reflected about the axis formed by the car's position and the final waypoint. {\bf (C):} In the case of lane changes, the guiding path is computed by a weighted average of the waypoints on the departure and destination lanes.}
    \label{fig:arcs}
\end{figure}

In order to perform trajectory planning, the ego-vehicle computes a set of waypoints along the center-line of its current lane at fixed time intervals, and these waypoints represent the expected positions for a planning horizon, $\tau$. We represent these waypoints as $\vec{w}_1 - \vec{w}_k$. Using its own position, the median point, and the final waypoint ($\vec{p}, \vec{w}_{\frac{k}{2}}, \vec{w}_k$), we  compute a circular arc on the road plane which sets the initial target speed and steering, $v'$ and $\phi'$ respectively, and acts as the guiding path for the next planning phase. We use circular arc approximations because they implicitly encode the radius of curvature needed for slip computation making it easy to check whether the dynamic constraints are violated. 

Absent discontinuities in the center-line of the lane, the guiding arcs exhibit first-order $C^1$ continuity. \Figref{fig:arcs}(a) demonstrates a guiding arc constructed for a sample lane. 

We constrain the arcs to lie within the first two quadrants of the circle that is represented by three waypoints. In cases when the vehicle's trajectory tracks away from the center of the lane, e.g.  during collision avoidance maneuvers, this constraint may be violated as shown in \figref{fig:arcs}(b). In such cases, the point $\vec{w}_{\frac{k}{2}}$ is reflected about the axis formed between $\vec{p}$ and $\vec{w}_k$ to correct the arc angle. In case of lane changing, waypoints are constructed from a weighted average of points sampled ahead on both the departure lane and the destination lane. \Figref{fig:arcs}(c) demonstrates a set of lane-change arcs.

Given a guiding path, a target steering, $\phi'$ is computed from \eqnref{eqn:motion:firstorder}. The radius of the arc, $r$, is substituted into \eqnref{eqn:latslip} to determine the maximum safe speed for the current road curvature. A target speed, $v'$, is computed from the minimum value of the current speed limit and the maximum safe speed. The target steering and speed form the basis of the control-obstacle exploration in the subsequent stage.


\subsubsection{Traffic Rules}

Traffic rules such as stopping at red lights are encoded in our algorithm. When choosing a target speed $v'$, the sensing system is referenced to determine if an intersection is being approached and whether the vehicle needs to stop at the intersection. In cases where the vehicle must stop, the edge of the intersection is used to compute a stopping point and $v'$ is set to the speed that will reach the stopping point at $\tau$ seconds. In case of stoplights, the green light signals $v'$ to return to its original value.

In the case of stops with continuous cross-traffic, the vehicle waits until the collision-avoidance algorithm indicates safety. This is accomplished by limiting the potential speed controls the vehicle may choose. When waiting for cross-traffic, the vehicle will stop until its guiding path is determined to be safe. 
In the case of all-way stops, the vehicle maintains a queue of vehicle arrival order, but defers to other drivers if they enter the intersection out of turn.

Although merges are not specifically encoded in transitions in the route plan, the vehicle is able to determine when merging is safe through the collision-avoidance and optimization stages of the algorithm. A merge is note determined safe and appropriate unless it provides collision-free guarantees and respects safety and comfort costs detailed in \secref{sec:optimization}.



\subsection{Collision Avoidance}
\label{sec:collision_avoidance}
We leverage the theory of \emph{Control Obstacles} for collision avoidance~\cite{Bareiss2015}. Control Obstacles construct constraints in the control space and are an extension of acceleration-velocity obstacles~\cite{VanDenBerg2011a}. For each neighbor of the ego-vehicle, $n$, we define the control obstacle for the neighbor as the union of all controls that could lead to collisions with the neighbor within the time horizon, $\tau$. Given t, where  $0 \leq t \leq \tau$, the relative position of the ego-vehicle and neighbor $\vec{p}_{en}$ must remain outside the Minkowski Sum given by the formulation, which is defined as
\begin{equation}
    \mc{O}_{en} = \mc{O}_n \oplus -\mc{O}_e.
\end{equation}
The complete derivation for control obstacles can be found in \cite{Bareiss2015}. 

In order to adapt to the autonomous vehicles, we modify the original control obstacle formulation \cite{Bareiss2015} in the following manner: (1) We do not assume reciprocity in collision avoidance and the ego-vehicle must take full responsibility for avoiding collisions; (2) We do not assume the control inputs of other vehicles are observable, which is consistent with the first point;  (3) We do not assume bounding discs for the neighboring entities, but rather a tight bounding rectangle. The Minkowski Sum for two convex polygons can be computed in linear time in the number of edges; (4) The new feasible control chosen does not correspond to  the control that minimizes the deviation from $v'$ and $\phi'$. Rather it is the control that minimizes the objective function defined in \secref{sec:optimization}.

The union of all control-obstacles and the set of dynamically infeasible controls form the boundary of the space of collision-free controls for the ego-vehicle. As long as a new control set is chosen from outside the union of the control obstacles, the ego-vehicle will be collision free for the next $\tau$ seconds. In \secref{sec:conclusion}, we detail how behavior prediction models can be incorporated to assume varying levels of reciprocity. This approach is conservative and it is possible that there may be no feasible solution. In that case, we reduce $\tau$ and search for a feasible solution.

\subsection{Trajectory Sampling}
\label{sec:sampling}
Computing the exact boundary of the control obstacle is computationally expensive. Moreover, depending on the choice of $A$ and $\Phi$, the boundary computation will typically not have an analytical solution. In order to ensure that the vehicle can plan within a specific time bound, we use a sampling strategy around $\phi'$ and $v'$ to determine a feasible control that the vehicle will adopt for the next $\tau$ seconds. Each sample is referred to as a \emph{candidate control} and represented as $u_c$. 

First, the closest collision-free velocity to $v'$ is determined where $\phi = \phi'$ by forward projection. This represents the largest speed the vehicle could take without deviating from the center-line of its lane and is always included in the set of candidates. Next, we compute evenly spaced samples around the point $(v', \phi')$ in the control space. We also choose a set of samples around the prior step solution, $\phi_{t-1}$ and $v_{t-1}$, which allows the vehicle to explore minor deviations in trajectory. Samples near the prior solution facilitate lane-keeping and within-lane avoidance maneuvers.

For each neighbor $n \in N$, we compute a set of states for that neighbor for the next $\tau$ seconds by forward integration of $q(X_n, \emptyset, t)$. We assume that the neighboring vehicle will follow its current lane at the current speed and acceleration during this time interval. Otherwise, the neighbor is assumed to move along its current velocity $\vec{v}_n$ with the current observed values of turning and acceleration, $\dot{\theta}$ and $\dot{v}_n$, respectively.

For each candidate control, $u_c$, we determine whether \eqnref{eqn:latslip} is violated by the candidate control inputs and immediately discard it if that is the case. If not, the sample points are computed at even time intervals along $0 \leq t \leq \tau$ by forward integration of $q(X_t, u_c, t)$. For each position in time, $\vec{p}_t$, we compute the relative position with each neighboring position at that time and determine if the relative position lies inside the Minkowski Sum. If so, we discard the candidate controls. After all the candidates are evaluated, the new control sequence is chosen by minimizing the objective function described in the subsequent section.


\subsection{New Trajectory Computation}
\label{sec:optimization}

\begin{figure*}
    \centering
    \includegraphics[scale=0.8]{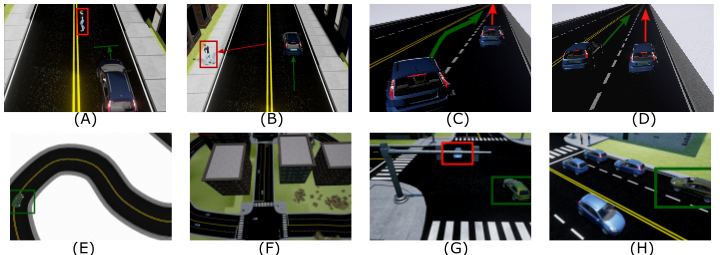}
    \caption{{\bf Results:} {\bf (A) and (B)}: The ego-vehicle is forced to stop as a pedestrian enters the roadway during the {\bf Jaywalking} benchmark due to the proximity costs. Once the pedestrian has moved away, the vehicle resumes its course. {\bf (C) and (D)}: The ego-vehicle approaches a slower moving vehicle from behind. The path and maneuver costs drive the ego-vehicle to plan a lane-change around the slower vehicle. The trajectory of the ego-vehicle is shown in green. {\bf (E)}: The Hatchback ego-vehicle during the {\bf S-Turns} benchmark. The vehicle plans the highest speed it can safely maintain during the tight turns. Each ego-vehicle plans a different safe speed based on their data-driven vehicle dynamics functions. {\bf (F)}: An overview of the {\bf Simulated City} benchmark. The ego-vehicle navigates amongst typical traffic to a set of randomly assigned destinations. {\bf (G)}: The ego-vehicle (outlined in green) yields to an oncoming vehicle (outlined in red) during the  {\bf Simulated City} benchmark. Once the vehicle clears the intersection, the ego-vehicle proceeds with a left turn. {\bf (H)}: The ego-vehicle (outlined in green) stops in traffic waiting for a stoplight to change during the {\bf Simulated City} benchmark.  }
    \label{fig:results}
\end{figure*}

Once a set of suitable control candidates has been computed, the vehicle selects the valid controls that minimize the 
following cost function at each sample point $i \in I$:
\begin{equation}
    C = \sum_{i=0}^{I}  c_{path}(i) + c_{cmft}(i) + c_{mnvr}(i) + c_{prox}(i).
\end{equation}
This function corresponds to producing paths which are comfortable for passengers, provide safe passing-distances from other vehicles, and respect the constraints of upcoming maneuvers the vehicle must perform. Each term consists of several cost evaluation functions, each with its own weight $e \in W$, which are described in the following sections. 

\subsubsection{Path Cost}
$c_{path}$ encodes costs associated with the vehicle's success at tracking its path and the global route. To compute this cost, we define two points. The target point, $\vec{p}_{tar}$, represents the relative position the vehicle would achieve following the spline defined by its guiding path exactly at $v'$. Given the final sample, we project the vehicle's expected position at the final sample point onto $\vec{p}_{tar}$, which we denote $\vec{p}_{I,tar}$. The path cost is given by:
\begin{eqnarray}
    c_{path} &=& c_{vel} + c_{drift} + c_{prog}\\
    c_{vel} &=& (v' - v)^2 \nonumber \\
    c_{drift} &=& ||\vec{p} - \vec{l}_p||^2 \nonumber \\
    c_{prog} &=& \frac{||p_{tar} - p_{I,tar}||}{||p_{tar}||}\nonumber 
\end{eqnarray}

$c_{vel}$ is the squared difference between desired speed and current speed and $c_{drift}$ is the squared distance between the center line of the vehicle's lane and its current position. If the path crosses a lane boundary, $c_{drift}$ is computed with respect to the new lane. $c_{prog}$ represents the vehicle's desire to maximally progress along its current path. Candidates which reduce progress with respect to the guiding path are penalized. These terms drive the vehicle to choose trajectories that maximally progress the ego-vehicle along its computed route between steps. $c_{prog}$ is only computed at the final sample point. 

\subsubsection{Comfort Costs}
Comfort costs are computed similar to  \cite{Ziegler2014b} and penalize motions which are uncomfortable for passengers in the vehicle. $c_{accel}$ penalizes large accelerations and decelerations. $c_{yawr}$ penalizes large heading changes and discourages sharp turning. The comfort costs are given as: 

\begin{eqnarray}
    c_{cmft} &=& c_{accel} + c_{yawr} \\ 
    c_{accel} &=& ||\dot{v}_i|| \nonumber \\
    c_{yawr} &=&  ||\dot{\theta}|| \nonumber
\end{eqnarray}

\subsubsection{Maneuver Costs}
The novel maneuvering cost function discourages lane-changes without excluding them and guides the vehicle to occupy the necessary lane for its next maneuver. The formulation is given as: 
\begin{eqnarray}
    c_{mnvr} &=& c_{lane} + c_{mdist} \\
    c_{lane} &=& 1 \cdot LaneChanged \nonumber\\
    c_{mdist} &=&  \frac{1}{\vec{p} - \vec{p}_{m}} \cdot WrongLane \nonumber
\end{eqnarray}
 $LaneChanged$ is a boolean variable representing whether a candidate path crosses a lane boundary. $\vec{p}_{m}$ is the position of the next maneuver change, e.g. the beginning of a right turn. This position is determined by the point of maneuver and starts in the desired lane for the maneuver. $WrongLane$ is a boolean that evaluates to true if the vehicle's lane does not match the lane for the next maneuver. If a candidate control is chosen where for some point $i \in I$, $LaneChanged$ evaluates to true, a lane change behavior is initiated in the finite state machine.

\subsubsection{Proximity Costs}
While the collision avoidance stage prevents the vehicle from colliding with neighbors, the proximity cost term is designed to prevent the vehicle from passing close to neighboring entities based on the identified type of the neighbor, $T_n$. This cost is represented as a cost distance term with exponential decay based on the relative positions of the ego-vehicle and its neighbor. 

\begin{eqnarray}
    c_{prox} = \sum_{n=0}^{N} d(N_{j}, \vec{p}) \\
    d(N_n, \vec{p}) = C_{type_{\mc{T}_n}}\cdot e^{-||\vec{p}_n - \vec{p}_e||}  \nonumber
\end{eqnarray}

$C_{type}$ is a per-type constant cost value. $C_{type}$ is larger for pedestrians and bicycles than for vehicles, and guides the ego-vehicle to pass those entities with greater distance.

\subsection{Data-driven Vehicle Dynamics Model}
\label{subsec:profiling}

In order to determine values for $A(v,u_t)$ and $\Phi(\phi, u_s)$, we use a data-driven approach to model the dynamics of the vehicle. For each ego-vehicle, data is collected by driving the vehicle from $v = 0$ to $v = v_{max}$ at the highest possible throttle without loss of traction.  Similarly, for braking, the vehicle is decelerated from $v = v_{max}$ to $v = 0$ using the highest braking effort possible without loss of traction. Data is collected at $60$Hz for these values: current speed, acceleration, and throttle/braking values. From these data, piecewise quadratic functions are constructed by least squares fitting to represent the maximum available acceleration and braking given the current vehicle state. These values also define thresholds for the control safety function $S(u, X)$.


We determine $\Phi(\phi,u_s)$ by fixing the vehicle's speed and collecting data for instantaneous changes to the steering angle for a given $u_s$. We construct a piecewise quadratic function by least-squares fitting to represent the vehicle's steering dynamics. Having the value of $\mu$ from the sensors, we determine the  maximum feasible speed for a given curvature from the centripetal force equation: 
\begin{equation}
\label{eqn:latslip}
    v = \sqrt{\mu r g}
\end{equation} where $r$ is the radius of curvature that is computed from \eqnref{eqn:motion:firstorder}. 
By substituting \eqnref{eqn:motion:firstorder} into \eqnref{eqn:latslip}, and the angular velocity formula $v = \omega \cdot r$, we can determine feasible steering for a given speed as
\begin{equation}
\label{eqn:latslip_steering}
    \phi = tan^{-1}(\frac{(L_f + L_r) \cdot \mu \cdot g)}{v^2}) .
\end{equation} Given the generated functions $S$, $A$, and $\Phi$, the future path of the vehicle can be evaluated quickly for planning future controls. 

\subsection{Control Input}

Once a new set of controls is chosen, they are input to the vehicle using a pair of PID controllers. One of the PID controllers drives the current speed to match the target speed. The second PID controller drives the current steering angle to match the target steering angle chosen by the optimization function. By limiting the choice of candidate controls to kinematically and dynamically feasible controls using our data-driven vehicle dynamics model, the PID controllers are sufficient to achieve the desired values.

%% file: experiment_definitions.tex
\section{Experimental Evaluation}
\label{sec:experiment:setup}

In this section, we detail the evaluation scenarios for our navigation algorithm. Each scenario is chosen to test different aspects of the algorithm including response time, safety, and handling different traffic situations.



\subsection{Ego-Vehicles}

To demonstrate the generality of our approach, we tested each experimental scenario on each of three vehicles. Vehicle 1, the hatchback, has a mass of $1365$~kg, a length of $3.8$m, and a maximum steering angle of $60^\circ$. Vehicle 2, the sports car, has a mass of $1750$~kg, a length of $4.6$m, and a maximum steering angle of $63^\circ$. Vehicle 3, the SUV, has a mass of $1866$~kg, a length of $4.8$m, and a maximum steering angle of $55^\circ$.

\subsection{Benchmarks}

We conducted a series of simulations with each vehicle representing a variety of the challenging traffic scenarios an ego-vehicle will face while navigating city roads and highways. 

\textbf{Passing a bicycle}: This scenario involves the ego-vehicle passing a bicycle on a four lane straight road. The vehicle should maintain a safe distance from the bicycle, changing lanes if possible to avoid the cyclist. We perform the evaluation twice, once featuring a vehicle in the adjacent lane preventing the vehicle from moving to avoid the cyclist without first adjusting its speed.

\textbf{Jaywalking Pedestrian}: This scenario features a pedestrian stepping into the road in front of the vehicle. The vehicle must react quickly to safely decelerate or stop to avoid the pedestrian.

\textbf{Sudden Stop at High Speed}: The vehicle must execute an emergency stop on a highway at high speeds when the vehicle in front of it stops suddenly. We evaluate this scenario in two conditions. First, we evaluate performance with no other traffic aside from the ego-vehicle and stopping vehicle. In this condition, swerving can be performed simply. Secondly, we evaluate this scenario with surrounding traffic, complicating any swerving maneuvers as the vehicle must account for nearby traffic.

\textbf{High Density Traffic Approaching a Turn}: This scenario features a four lane road with the ego-vehicle starting in a heavily congested outer lane. The ego-vehicle must make a turn at a stoplight ahead in the outer lane. To make optimal progress, the ego-vehicle must execute a lane change to the inner lane, but must return to the outer lane with sufficient time to execute the turn. 

\textbf{Car Suddenly entering Roadway}: This scenario demonstrates the ego-vehicle traveling along a straight road at constant speed when a vehicle suddenly enters the roadway ahead of the vehicle and blocks the vehicle's path. The vehicle must decelerate and swerve to avoid colliding with the blocking vehicle. We demonstrate this scenario with the ego-vehicle travelling at 10, 30, and 50 mph and with the blocking vehicle obstructing either the right lane or both lanes.

\textbf{S-turns}: We demonstrate the ego-vehicle navigating a set of tight alternating turns, or S turns. Each ego-vehicle computes a different safe speed depending on the specific kinematic and dynamic limits of the vehicle. 

\textbf{Simulated City}: We demonstrate the ego-vehicle navigating to several key points in a small simulated city. The vehicle must execute lane changes to perform various turns as it obeys traffic laws and navigates to its goals. The vehicle encounters bicycles, pedestrians, and other vehicles as it navigates to its waypoints.

%% file: experiment_results.tex
\section{Benchmark results}
\label{sec:experiment:results}

We evaluated our navigation algorithm in these simulated scenarios. The algorithm can avoid tens of vehicles at interactive rates. As expected, the sports-car and the hatchback were able to maintain their preferred speeds more effectively in turns, whereas our SUV was forced to reduce speed. Each of the vehicles was able to pass other vehicles, pedestrians, and bicycles safely. In 
{\bf Car Suddenly entering Roadway} scenario, we observed a greater tendency to swerve. 
We did not observe the ego-vehicle colliding with any of the simulated vehicles in traffic.

\Figref{fig:results} details some of the interesting behaviors we observed while testing our navigation algorithm.  As expected, the ego-vehicle utilizes lane-changes to pass slower vehicles when no traffic is imposing. In traffic, the ego-vehicle slows down until it is safe to pass in the adjoining lane.
When interacting with pedestrians, the high proximity cost discourages the vehicle from changing lanes as the pedestrian passes, and the vehicle instead waits until the pedestrian has moved considerably.

\subsection{Timing Results}

\begin{figure}
  \begin{center}
    \includegraphics[width=.42\textwidth]{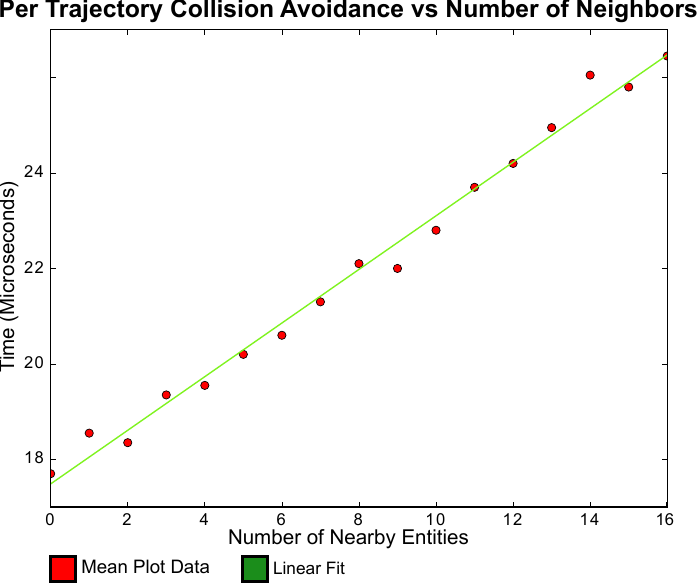}
  \end{center}
  \vspace{-16pt}
  \caption{ {\bf Collision Avoidance Timing:} We detail the relationship between the cost of collision checking for a trajectory and the number of nearby entities considered. We observe a linear relationship between collision checking and number of neighbors.}
  \label{fig:timing::CA}
\end{figure}

We collected data from a simulation designed to gradually increase the density of other vehicles encountered by our car. \Figref{fig:timing::CA} demonstrates the relationship between collision avoidance cost for a specific trajectory sample and the number of nearby vehicles and entities. We observe the cost of collision avoidance grows linearly in the number of neighbors. 

\Figref{fig:timing::CostEval} demonstrates the relationship between the number of trajectories sampled and the computational expense of optimization evaluation. The observed relationship is linear with greater variance than that of collision avoidance computation. The overall computation time for a typical navigation update including guiding path computation, control-obstacle sampling, collision-avoidance and cost evaluation is on the order of milliseconds, typically between 1 and 2 milliseconds. This suggests that the cost of the algorithm is dominated by the optimization time. 

%% file: conclusion.tex
\section{CONCLUSION AND LIMITATIONS}
\label{sec:conclusion}

\begin{figure}
	\begin{center}
		\includegraphics[width=.42\textwidth]{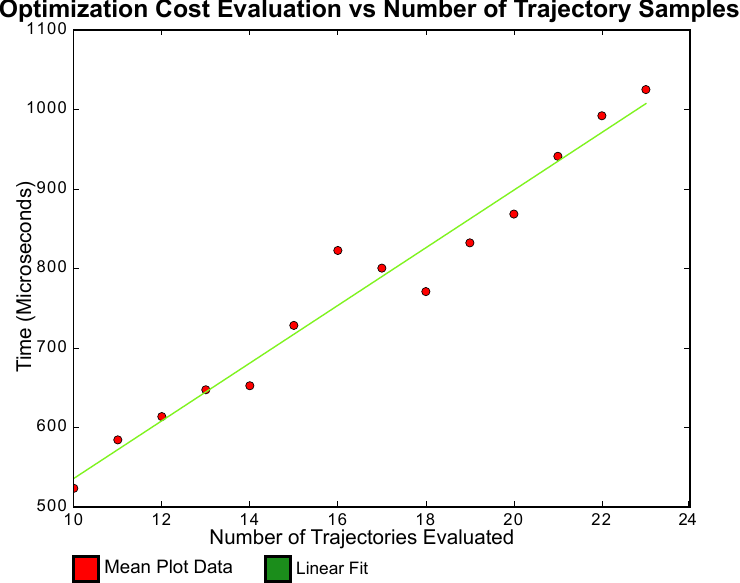}
	\end{center}
	\vspace{-16pt}
	\caption{ {\bf Cost Function Evaluation Timing:} The computational cost of computing the optimal trajectory for the vehicle varies linearly with the number of collision-free trajectories evaluated. }
	\label{fig:timing::CostEval}
\end{figure}

We present, AutonoVi, a navigation algorithm for autonomous vehicles. Our approach uses a data-driven vehicle dynamics model and optimization-based maneuver planning to compute safe, collision free trajectories with dynamic lane changes under typical traffic conditions. We have demonstrated our algorithm on a varied set of vehicles under varying dense and sparse traffic conditions with pedestrians and cyclists. We have also demonstrated that our vehicles follow traffic laws, and utilize both braking and steering simultaneously when avoiding collisions. We highlight many benefits over prior methods in our simulations.

Our approach has some limitations. First, though our introduction of the data-driven dynamics functions $A$, $\Phi$, and $S$ generalize to arbitrary levels of underlying dynamics complexity, our current approach requires computing new vehicle dynamics functions for different values of $\mu$. We will address this limitation in future work by learning a transfer function between various road frictions to produce more general data-driven vehicle dynamics functions. In addition, we have assumed perfect sensing in the current technique and it would be useful to take into account sensing errors and uncertainty in our approach. 
These could be based on relying on predictive behavior models to overcome imperfect state estimations for neighboring entities \cite{Galceran2015,Sun2014}. With prediction, our control obstacles could anticipate levels of reciprocity from predictable vehicles. We will also explore whether the use of circular arcs may is appropriate for vehicles with substantially different geometries, such as trucks pulling large trailers. It is unclear if our kinematic models and data-driven profiles will navigate large vehicles safely. We will also like to incorporate real-world driving patterns and cultural norms to improve our navigation algorithm. This will include choosing optimal weights for the navigation algorithm using available training data.

